\documentclass[cameraready]{Interspeech}

\title{AV-SyncBench: Decoupled Benchmarking of Temporal and Semantic Audio-Visual Synchronization}

\author[affiliation={2}]{Tianhong}{Zhou}
\author[affiliation={1}]{Mingyang}{Han}
\author[affiliation={1}]{Boyu}{Li}
\author[affiliation={2}]{Yuxuan}{Jiang}
\author[affiliation={3}]{Jiaxin}{Ye}
\author[affiliation={1}]{Dongxiao}{Wang}
\author[affiliation={1}]{Haoxiang}{Shi}
\author[affiliation={1}]{Kunpeng}{Wang}
\author[affiliation={1}, correspondingauthor]{Jun}{Song}
\author[affiliation={1}, correspondingauthor]{Cheng}{Yu}
\author[affiliation={1}, correspondingauthor]{Bo}{Zheng}

\address{
    $^1$ Alibaba Group, China \\
    $^2$ Tsinghua University, Beijing, China \\
    $^3$ Fudan University, Shanghai, China
}

\email{
zth24@mails.tsinghua.edu.cn,
jsong.sj@alibaba-inc.com
}


\usepackage{comment}
\usepackage{booktabs}
\usepackage{multirow}
\usepackage{tabularx}

\begin{document}

\maketitle

\begin{abstract}
Audio–visual feature extraction is a fundamental component of multimodal understanding and generation tasks. However, existing evaluation protocols for feature extraction models exhibit dimensional bias, typically focusing on either semantic matching or temporal offset detection. Moreover, their data construction remains coupled, preventing independent assessment of temporal and semantic consistency. We propose AV-SyncBench, the first benchmark to fully separate temporal and semantic evaluation for audio–visual synchronization. Built from in-the-wild videos, it spans Voice, Music, and Sound across 10 scenarios and 5 challenge tasks. Data are automatically filtered and manually verified to ensure on-screen sound sources. The benchmark contains 3,269 videos and 38,390 samples, and we evaluate five representative models to quantify feature quality for alignment and downstream tasks. The code and dataset are available at: \url{https://fgt7t6g.github.io/AV-SyncBench}.
\end{abstract}

\section{Introduction}

The inherent audio-visual alignment of video data provides an exceptionally rich supervisory signal for self-supervised learning~\cite{arandjelovic2017look,owens2018multisensory}. High-quality features endowed with precise audio-visual alignment are not only foundational for multimodal understanding tasks such as audio-visual event classification~\cite{tian2018ave} and sound source localization~\cite{zhao2018soundofpixels}, but also an indispensable prerequisite for emerging audio-visual generation tasks like Video-to-Audio (V2A), Audio-to-Video (A2V), and Text-to-Audio-Video (T2AV)~\cite{luo2023difffoley,cheng2025mmaudio}. Recent timing-controllable audio generation methods further demonstrate the importance of preserving both semantic content and temporal structure~\cite{jiang2025freeaudio,jiang2025controlaudio}.

Specifically, existing audio–visual understanding models (e.g., VATT~\cite{akbari2021vatt}, Qwen3-Omni~\cite{xu2025qwen3omni}) rely on both fine-grained temporal correspondence and high-level semantic alignment. Moreover, some existing audio-visual generation works like MMAudio~\cite{cheng2025mmaudio} and Kling-Foley~\cite{wang2025klingfoley} have significantly enhanced synchronization effects by incorporating features extracted from dedicated audio-visual alignment modules. Concurrently, various audio-visual generation tasks (e.g., V2A, A2V, T2AV) critically rely on high-quality, highly synchronous audio-visual training data. Effectively sifting through vast, heterogeneous datasets to identify suitable training samples necessitates a robust audio-visual alignment feature extractor~\cite{chen2021avsync}. For these challenging tasks, an audio-visual feature extractor must possess two core capabilities: effectively capturing high-level cross-modal \textbf{Semantic Alignment} and accurately preserving \textbf{fine-grained temporal correspondence}.

However, current evaluation paradigms for audio-visual feature extraction models suffer from significant limitations, often conflating these two critical abilities or prioritizing one at the expense of the other~\cite{chen2021avsync}. Many mainstream multimodal foundation models (e.g., CLAP~\cite{elizalde2023clap}, ImageBind~\cite{girdhar2023imagebind}, CAV-MAE~\cite{gong2023cavmae}) are primarily assessed via cross-modal retrieval tasks. Although these benchmarks excel at measuring global semantic recall and matching accuracy, they generally exhibit limited sensitivity to subtle temporal misalignments~\cite{iashin2022sparsesync}. Furthermore, studies specifically investigating audio-visual synchronization (e.g., Synchformer~\cite{iashin2024synchformer}, SparseSync~\cite{iashin2022sparsesync}) predominantly emphasize the detection accuracy of temporal offsets, with a primary focus on evaluating temporal alignment capabilities; nevertheless, these approaches often neglect the features' robustness to semantic-level perturbations (e.g., timbre or sound-source changes under fixed timing)~\cite{engel2020ddsp}.

For a truly general-purpose audio-visual feature extractor, both semantic alignment and temporal information preservation are equally crucial, jointly defining the performance ceiling for downstream audio-visual understanding and synchronous generation tasks. Regrettably, the academic community currently lacks a benchmark capable of entirely decoupling ``temporal consistency'' and ``semantic consistency,'' and systematically and quantitatively evaluating these two distinct dimensions of capability.

To address this critical gap, we introduce \textbf{AV-SyncBench}, a pioneering, decoupled evaluation framework designed to independently diagnose the true capabilities of feature extractors in maintaining fine-grained temporal alignment and semantic consistency.

\begin{figure*}[t]
    \centering
    \includegraphics[width=\textwidth]{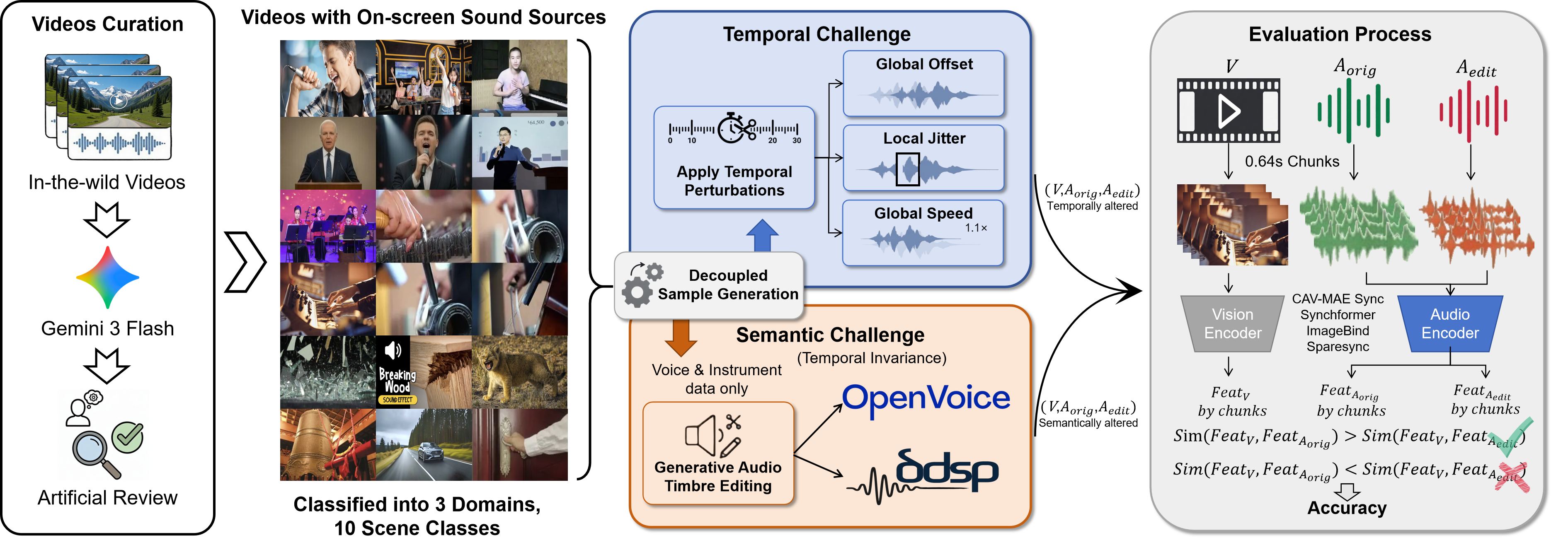}
    \caption{Overview of the AV-SyncBench decoupled evaluation framework. The benchmark is built from curated in-the-wild videos and generates two independent challenge sets: temporal challenges (global offset, local jitter, global speed change) and semantic challenges (timbre replacement, sound-source replacement) under fixed timing. Videos and original/edited audios are segmented into fixed-length chunks for feature extraction, and synchronization quality is quantified by comparing diagonal cosine similarities and reporting binary accuracy.}
    \label{fig:avsyncbench-pipeline}
\end{figure*}

The main contributions of this paper are summarized as follows:
\begin{itemize}
\item We propose AV-SyncBench, the first benchmark that explicitly decouples temporal perception and semantic consistency in audio-visual feature extraction into distinct challenging tasks.
\item We design a robust data collection pipeline, utilizing *in-the-wild* videos to construct the dataset, thereby maximizing the reduction of data leakage risks prevalent with traditional training sets (e.g., AudioSet~\cite{gemmeke2017audioset}, VGGSound~\cite{chen2020vggsound}).
\item We introduce novel semantic challenge tasks, built upon advanced generative editing techniques (OpenVoice~\cite{qin2024openvoice,myshell2024openvoicev2} and DDSP~\cite{engel2020ddsp}), which can alter semantics (e.g., timbre and instrument type) without disrupting the original temporal rhythm (thus preserving temporal information).
\item We systematically evaluate 5 state-of-the-art audio-visual feature extraction models (including Synchformer, CAV-MAE Sync~\cite{araujo2025cavmaesync}, ImageBind, etc.), providing an comprehensive analysis of their specific strengths and limitations within semantically and temporally decoupled scenarios.
\end{itemize}

\section{Methodology}

We propose a comprehensive benchmarking pipeline to systematically evaluate the synchronization capabilities of audio–visual feature extraction models by explicitly integrating decoupled assessments of temporal and semantic consistency, thereby addressing a critical gap in existing feature evaluation paradigms. The framework defines audio–visual synchronization performance along two key dimensions: temporal consistency perception and semantic consistency perception. The first dimension measures a model’s sensitivity to temporal misalignment, focusing on its ability to detect global offsets, local jitter, and global speed variations between visual and audio streams. The second dimension evaluates the model’s capacity to discriminate core semantic attributes, requiring it to correctly identify changes such as timbre replacement or sound-source substitution while strictly preserving the original physical timing and rhythm. All evaluations are conducted on rigorously curated in-the-wild videos to effectively mitigate potential data leakage from common pretraining datasets, ensuring the objectivity, generalizability, and reliability of the benchmark results.

\subsection{The AV-SyncBench Pipeline}

Figure~\ref{fig:avsyncbench-pipeline} illustrates the overall pipeline of AV-SyncBench. The framework is designed to systematically evaluate audio–visual feature extraction models by explicitly decoupling temporal consistency and semantic consistency under a unified protocol.

We first construct and curate a collection of in-the-wild videos spanning three major domains: Voice, Music, and Sound. Each video is annotated with one of 10 scenario labels and undergoes manual verification to ensure that the primary sound source is visually observable within the frame, thereby establishing reliable audio–visual correspondence and mitigating potential data leakage from commonly used pretraining datasets.

Given an original pair $(V, A_{\text{orig}})$, we generate two independent challenge tracks. In the \textbf{temporal challenge}, we preserve semantic content while introducing controlled temporal perturbations, including global offset, local jitter, and global speed change, to evaluate sensitivity to misalignment. In the \textbf{semantic challenge}, we enforce temporal invariance and modify only semantic attributes (e.g., timbre or sound-source replacement), producing an edited audio $A_{\text{edit}}$ that shares identical timing with $A_{\text{orig}}$ but differs semantically.

For evaluation, videos and audios are segmented into fixed-length 0.64-second chunks (small, non-overlapping video segments), and visual and audio embeddings are extracted independently. Synchronization quality is measured via diagonal cosine similarity, and performance is reported as binary accuracy by comparing original pairs against perturbed or edited pairs.

This decoupled design enables direct diagnosis of whether a model captures temporal alignment, semantic alignment, or both.

\subsection{Dataset Construction}

To construct a dataset that faithfully reflects audio–visual synchronization ability, we collect in-the-wild videos from public online platforms and implement a strict data curation and variable control pipeline to ensure reliable cross-modal correspondence and objective evaluation. Quality control follows a two-stage process: Gemini 3 Flash~\cite{doshi2025gemini3flash} is first used to remove samples with off-screen sound sources or obvious audio–visual mismatches. The remaining samples are then manually reviewed by five human annotators, where each video clip is independently examined by at least three annotators to verify that the primary sound source is visible on screen and temporally aligned with the corresponding visual events, while clips with poor audio quality, excessive noise, or ambiguous semantics are removed. After trimming and normalization, we retain 3,269 high-quality video clips with durations ranging from 3 to 13 seconds. The dataset spans three domains—Voice, Music, and Sound—covering 10 scenario types, and each video is annotated with a scenario label for category-wise analysis.

Based on these curated videos, we construct five decoupled challenge tasks through controlled manipulation of temporal and semantic variables. Temporal challenges modify only timing while preserving semantic content. Global Offset shifts the entire audio track by 50–500 ms (five discrete levels, with zero-padding applied). Local Jitter introduces random local shifts of 30–700 ms (categorized as mild: 30–70 ms, moderate: 150–250 ms, and severe: 400–600 ms), with the shift applied to randomly selected positions. One 2-second window is perturbed for 5-second clips and two 2-second windows for 10-second clips. Global Speed Change adjusts the playback speed of the entire audio-visual track within 0.8×–1.25× using 10 discrete levels. To ensure consistent duration, all tracks are truncated to the shortest length.

Semantic challenges strictly preserve the original temporal structure while altering semantic attributes. In voice scenarios, OpenVoice V2 is used for timbre replacement, with reference samples covering gender and age categories (child, youth, adult, elderly). In instrument scenarios, a pretrained DDSP timbre transfer model performs instrument-specific conversion while preserving rhythm and pitch contour.

Through this variable-isolated construction, we obtain a benchmark covering five challenge types, enabling independent control of temporal and semantic factors and supporting decoupled diagnosis of audio–visual feature extraction models. Detailed dataset statistics are summarized in Table~\ref{tab:avsyncbench_stats}.

\begin{table}[t]
\centering
\scriptsize
\setlength{\tabcolsep}{3pt}
\caption{AV-SyncBench dataset statistics.}
\label{tab:avsyncbench_stats}
\begin{tabular}{l r l r l r}
\toprule
\multicolumn{2}{c}{\textbf{(A) Scenario}} &
\multicolumn{2}{c}{\textbf{(B) Duration (s)}} &
\multicolumn{2}{c}{\textbf{(C) Tasks}} \\
\cmidrule(lr){1-2} \cmidrule(lr){3-4} \cmidrule(lr){5-6}
Scenario & \ Videos & Range & \ Videos & Task & \ Samples / \ Videos \\
\midrule
Action & 248 & 3--5 & 1,586 & Global Offset & 15,000 / 1,500 \\
Animal Sound & 274 & 5--7 & 346 & Global Speed & 15,000 / 1,500 \\
Object Sound & 319 & 7--9 & 1,253 & Local Jitter & 7,569 / 1,500 \\
Ambient Sound & 320 & 9--11 & 2,064 & Voice Timbre & 592 / 323 \\
Group Vocal. & 392 & 11--13 & 995 & Instr. Timbre & 229 / 229 \\
Single Speaker & 448 & & & \textbf{Temporal} & \textbf{37,569 / 2,717} \\
Dialogue & 319 & & & \textbf{Semantic} & \textbf{821 / 552} \\
Singing & 158 & & &  &  \\
Single Instr. & 423 & & & & \\
Ensemble & 366 & & & & \\
\midrule
\textbf{Total} & \textbf{3,269} &  &  & \textbf{Total} & \textbf{38,390 / 3,269}\\
\bottomrule
\end{tabular}
\end{table}

\subsection{Metrics}

We evaluate synchronization under a unified pairwise comparison protocol. 
For contrastive representation models, videos and audios are segmented into non-overlapping 0.64-second chunks, and visual embeddings $v_i$ and audio embeddings $a_i$ are extracted independently. Synchronization strength is defined as the mean diagonal similarity:

\[
S = \frac{1}{N} \sum_{i=1}^{N} \mathrm{sim}(v_i, a_i),
\]

where $\mathrm{sim}(\cdot,\cdot)$ denotes cosine similarity for contrastive models. 
For offset-classification models, which output a probability distribution over discrete temporal offsets, we use the predicted probability of zero offset, $p(\Delta = 0)$, as the synchronization score.

For both model types, a prediction is considered correct if the original audio--video pair receives a higher synchronization score than its perturbed or edited counterpart. 
Final performance is reported as binary classification accuracy.

\section{Experiment}

\subsection{Setup}

All experiments are conducted on two NVIDIA H20 GPUs, with each job allocated 4 vCPUs (Intel Xeon Platinum 8469C). All models are evaluated using their officially released codebases and pretrained checkpoints, without any additional training or fine-tuning. We benchmark five representative audio–visual models, including Synchformer(VGGSound version), SparseSync(VGGSound-Sparse version), ImageBind, CAV-MAE(Scale++ version), and CAV-MAE-Sync. For each model, we strictly follow the preprocessing pipeline, input resolution, and inference configurations specified in the official repositories to ensure reproducibility and fairness.

To maintain evaluation consistency, all videos are decoded at 25 FPS and audio is resampled to 16 kHz. Moreover, all models are evaluated under the same segmentation protocol (0.64-second non-overlapping chunks) and the same pairwise decision criterion.
\begin{table}[t]
\centering
\small
\setlength{\tabcolsep}{3pt}
\caption{Accuracy under different settings for three temporal challenges: Global Offset, Local Jitter, and Global Speed Change. Best results are \textbf{bold} and second-best are \underline{underlined}.}
\label{tab:temporal_breakdown}

\resizebox{0.99\columnwidth}{!}{
\begin{tabular}{@{}lccccc@{}}
\toprule
\textbf{Setting} & \textbf{Synchformer} & \textbf{ImageBind} & \textbf{CAV-MAE-Sync} & \textbf{SparseSync} & \textbf{CAV-MAE} \\
\midrule

\multicolumn{6}{l}{\textit{Global Offset (ms)}} \\
50  & \underline{0.510} & 0.493 & 0.495 & \textbf{0.518} & 0.495 \\
100 & \textbf{0.541} & 0.485 & 0.486 & \underline{0.514} & 0.503 \\
200 & \textbf{0.582} & 0.491 & 0.492 & \underline{0.561} & 0.501 \\
300 & \textbf{0.622} & 0.512 & 0.511 & \underline{0.602} & 0.476 \\
500 & \textbf{0.662} & 0.542 & 0.517 & \underline{0.648} & 0.557 \\
Overall & \textbf{0.583} & 0.505 & 0.500 & \underline{0.569} & 0.506 \\

\midrule
\multicolumn{6}{l}{\textit{Local Jitter (ms range)}} \\
L1 (30--70)   & 0.639 & 0.572 & 0.662 & \textbf{0.729} & \underline{0.666} \\
L2 (150--250) & 0.723 & 0.593 & 0.639 & \underline{0.729} & \textbf{0.806} \\
L3 (400--600) & \underline{0.804} & 0.690 & 0.608 & 0.717 & \textbf{0.832} \\
Overall & 0.722 & 0.618 & 0.636 & \underline{0.725} & \textbf{0.768} \\

\midrule
\multicolumn{6}{l}{\textit{Global Speed Change}} \\
0.80$\times$ & 0.610 & \underline{0.792} & 0.506 & 0.615 & \textbf{0.846} \\
0.83$\times$ & 0.605 & \underline{0.763} & 0.485 & 0.592 & \textbf{0.847} \\
0.87$\times$ & 0.616 & \underline{0.722} & 0.527 & 0.629 & \textbf{0.836} \\
0.91$\times$ & 0.602 & \underline{0.666} & 0.482 & 0.638 & \textbf{0.795} \\
0.95$\times$ & 0.572 & \underline{0.633} & 0.566 & 0.577 & \textbf{0.707} \\
1.05$\times$ & \underline{0.588} & 0.498 & 0.467 & \textbf{0.681} & 0.514 \\
1.10$\times$ & \underline{0.614} & 0.491 & 0.455 & \textbf{0.795} & 0.559 \\
1.15$\times$ & \underline{0.611} & 0.486 & 0.497 & \textbf{0.846} & 0.565 \\
1.20$\times$ & \underline{0.629} & 0.483 & 0.438 & \textbf{0.850} & 0.559 \\
1.25$\times$ & \underline{0.628} & 0.486 & 0.441 & \textbf{0.848} & 0.543 \\
Overall & 0.607 & 0.602 & 0.486 & \textbf{0.707} & \underline{0.677} \\

\bottomrule
\end{tabular}
}
\end{table}

\begin{table*}[t]
\centering
\small
\setlength{\tabcolsep}{4pt}
\caption{Average accuracy across each category. Best results in each category are in \textbf{bold}, and second-best results are underlined.}
\label{tab:temporal_category_avg}

\begin{tabular}{lccccccccccccc}
\toprule
\textbf{Model} 
& \multicolumn{2}{c}{\textbf{Music}}
& \multicolumn{4}{c}{\textbf{Sound}}
& \multicolumn{4}{c}{\textbf{Voice}}
& \multicolumn{3}{c}{\textbf{Avg}} \\
\cmidrule(lr){2-3}
\cmidrule(lr){4-7}
\cmidrule(lr){8-11}
\cmidrule(lr){12-14}

& Inst & Ens
& Obj & Act & Amb & Ani
& Spk & Grp & Dial & Sing
& Music & Sound & Voice \\
\midrule
Synchformer 
& \underline{0.656} & \underline{0.646}
& 0.613 & \underline{0.655} & 0.638 & \underline{0.621}
& \underline{0.683} & 0.603 & 0.617 & \textbf{0.662}
& \underline{0.651} & 0.632 & 0.641 \\

ImageBind 
& 0.581 & 0.542
& 0.572 & 0.599 & 0.578 & 0.571
& 0.605 & 0.571 & 0.554 & 0.592
& 0.562 & 0.580 & 0.580 \\

CAV-MAE-Sync 
& 0.591 & 0.475
& 0.595 & 0.596 & 0.470 & 0.506
& 0.601 & 0.540 & 0.558 & 0.414
& 0.533 & 0.542 & 0.528 \\

SparseSync
& 0.650 & \textbf{0.701}
& \textbf{0.673} & \textbf{0.662} & \underline{0.644} & \textbf{0.640}
& \textbf{0.695} & \underline{0.680} & \textbf{0.659} & \underline{0.658}
& \textbf{0.676} & \textbf{0.655} & \textbf{0.673} \\

CAV-MAE
& \textbf{0.666} & 0.630
& \underline{0.662} & 0.636 & \textbf{0.691} & 0.583
& 0.636 & \textbf{0.683} & \underline{0.652} & 0.651
& 0.648 & \underline{0.643} & \underline{0.656} \\

\bottomrule
\end{tabular}

\end{table*}

\begin{table*}[t]
\centering
\footnotesize
\setlength{\tabcolsep}{5pt}
\renewcommand{\arraystretch}{0.85}
\setlength{\aboverulesep}{1pt}
\setlength{\belowrulesep}{1pt}
\setlength{\abovecaptionskip}{2pt}
\setlength{\belowcaptionskip}{2pt}
\caption{Timbre editing accuracy across models (overall and by category). Best results are \textbf{bold} and second-best are \underline{underlined}.}
\label{tab:timbre_all}
\begin{tabular}{@{}lccccc@{}}
\toprule
\textbf{Category} & \textbf{Synchformer} & \textbf{ImageBind} & \textbf{CAV-MAE-Sync} & \textbf{SparseSync} & \textbf{CAV-MAE} \\
\midrule
\textbf{Overall} & 0.787 & \textbf{0.859} & 0.628 & 0.485 & \underline{0.826} \\
\midrule
\multicolumn{6}{c}{\textit{Voice-related}} \\
Single Speaker & 0.734 & \textbf{0.933} & 0.510 & 0.436 & \underline{0.828} \\
Multi Speaker & 0.823 & \textbf{0.957} & 0.421 & 0.531 & \underline{0.824} \\
Singing & 0.693 & \textbf{0.872} & 0.541 & 0.403 & \underline{0.761} \\
Voice Avg & 0.750 & \textbf{0.935} & 0.491 & 0.457 & \underline{0.804} \\
\midrule
\multicolumn{6}{c}{\textit{Music-related}} \\
Single Instrument & 0.820 & 0.787 & \underline{0.855} & 0.574 & \textbf{0.899} \\
Ensemble & \textbf{0.864} & 0.702 & 0.815 & 0.482 & \underline{0.819} \\
Instrument Avg & \underline{0.842} & 0.745 & 0.835 & 0.528 & \textbf{0.859} \\
\bottomrule
\end{tabular}
\vspace{-1.5mm}
\end{table*}

\subsection{Results}

We analyze the results by aggregating model performance from three complementary perspectives: perturbation intensity, category, and semantic editing tasks. Due to space constraints, we mainly present representative results for key tasks and category settings, and summarize performance across perturbation intensities and categories to highlight differences among models in temporal and semantic consistency.

Table~\ref{tab:temporal_breakdown} shows model performance under different temporal perturbations. As the global offset increases from 50 ms to 500 ms, accuracy generally improves, and a similar trend is observed for local jitter and global speed change, indicating that larger perturbations are easier to detect. However, all models achieve relatively low scores on the global offset task, suggesting that fine-grained temporal misalignment remains challenging, as the audio features themselves are naturally continuous, making it difficult for the models to identify and align subtle temporal deviations. Synchformer performs best in the offset setting, showing strong sensitivity to temporal shifts, while SparseSync also performs well due to its offset-focused training. In contrast, CAV-MAE shows near-random performance on the offset task, indicating that its representations capture coarser temporal structures rather than precise temporal alignment.

Table~\ref{tab:temporal_category_avg} further analyzes the category-level average accuracy of different models across the three temporal perturbation tasks. From the category distribution, different models exhibit clear performance biases across scenarios. Taking Synchformer as an example, it performs particularly well in scenarios with clear sound sources, such as single-speaker narration and singing, but its performance drops noticeably in more complex settings such as group vocalization, dialogue, and object collision events. A similar trend can also be observed in other categories: for instance, single-instrument scenarios generally achieve higher accuracy than ensemble performances. Overall, when the sound source is explicit and closely aligned with visible actions, models are more capable of capturing temporal consistency. In contrast, in multi-source or complex interaction scenarios, where audio sources are mixed and visual cues become less clear, model performance tends to degrade.

Table~\ref{tab:timbre_all} presents the results of the timbre editing task with unchanged temporal structure. ImageBind achieves the highest accuracy, excelling in voice-related scenarios, indicating its effectiveness in distinguishing semantic changes in timbre. SparseSync, performing close to random, struggles to identify semantic differences when only timbre changes. CAV-MAE and CAV-MAE Sync perform well in music scenarios, showing strong instrument timbre recognition. Synchformer shows balanced performance, with good accuracy across most categories.

Overall, these results further indicate a clear decoupling between temporal consistency and semantic consistency in model capabilities. For instance, ImageBind performs strongly in semantic discrimination but relatively weakly in temporal alignment tasks, while SparseSync exhibits the opposite trend.

\section{Limitations}
Although AV-SyncBench provides a framework for decoupled evaluation of temporal and semantic consistency, several limitations remain. First, the semantic editing tasks rely on generative methods such as DDSP and OpenVoice V2. While these methods preserve temporal structure, differences in generation mechanisms may introduce subtle acoustic variations beyond pure timbre changes. As a result, the edited scenarios produced by different editing pipelines are not strictly comparable. Second, the current semantic edits mainly focus on speech and music scenarios, while controllable replacement of object-related sounds (e.g., collision or environmental effects) remains limited. In addition, the benchmark is primarily constructed from video segments shorter than 13 seconds and does not yet cover longer temporal contexts or more complex multi-source interactions. Future work could address these limitations by adopting more precise audio editing techniques and expanding the diversity and scale of evaluation scenarios.

\section{Conclusion}

This paper introduces AV-SyncBench, a benchmark designed to decouple the evaluation of temporal consistency and semantic consistency in audio–visual models. By constructing temporal perturbation and semantic editing tasks, the benchmark systematically evaluates existing audio–visual feature extractors along two dimensions: fine-grained temporal alignment and semantic discrimination. Under a unified evaluation protocol, we compare several representative models and provide clearer insights into the capability structure of current audio–visual representations.

Our findings reveal a mismatch in the training objectives of current audio–visual models. Existing approaches typically emphasize either temporal synchronization (e.g., offset detection) or semantic association (e.g., cross-modal contrastive learning), while rarely modeling both simultaneously. This bias leads to noticeable performance differences across tasks, suggesting that future models should adopt more unified learning mechanisms that jointly capture temporal structure and semantic alignment.

\newpage

\section{Generative AI Use Disclosure}
Generative AI tools were used exclusively to assist in editing and improving the manuscript’s language for better clarity and flow. These tools did not contribute to the creation of any scientific content, data, or conclusions. All authors thoroughly reviewed and revised the final manuscript and are fully responsible for its accuracy and integrity.

\bibliographystyle{IEEEtran}
\bibliography{references}

\end{document}